\documentclass[conference]{IEEEtran}
\IEEEoverridecommandlockouts
\usepackage{cite}
\usepackage{amsmath,amssymb,amsfonts}
\usepackage{graphicx}
\graphicspath{{figures/}{../figures/}{./}}
\usepackage{textcomp}
\usepackage{xcolor}
\usepackage{booktabs}
\usepackage{multirow}
\usepackage{array}
\usepackage{enumitem}
\usepackage{dblfloatfix}
\usepackage{float}
\usepackage[T1]{fontenc}
\usepackage[utf8]{inputenc}
\usepackage{tikz}
\usetikzlibrary{positioning,arrows.meta,calc}
\usepackage[hidelinks]{hyperref}
\def\BibTeX{{\rm B\kern-.05em{\sc i\kern-.025em b}\kern-.08em
    T\kern-.1667em\lower.7ex\hbox{E}\kern-.125emX}}

\title{Latent Space Probing for Adult Content Detection in Video Generative Models}

\author{
\IEEEauthorblockN{Alizishaan Khatri}
\IEEEauthorblockA{\textit{Wrynx Inc.}\\
\texttt{research@wrynx.com}}
\and
\IEEEauthorblockN{Chiquita Prabhu}
\IEEEauthorblockA{\textit{Independent Researcher}}
}

\begin{document}
\maketitle

\begin{abstract}
The rapid proliferation of AI-powered video generation systems has introduced significant challenges in content moderation, particularly with respect to adult and sexually explicit material. Existing  detection methods operate on either prompts or  decoded pixel-space outputs. Therefore, both approaches are blind to the rich internal representations formed during generation. In this paper, we propose a novel \emph{latent space probing} framework that intercepts the denoised latent representations produced by the CogVideoX video diffusion model during inference and attaches lightweight classifiers to perform real-time adult content detection. To support this work, we construct a large-scale binary dataset of 11039 ten-second video clips (5086 violating, 5953 non-violating) sourced from adult websites and YouTube respectively. We introduce two lightweight probing classifier architectures. We train and evaluate it on the dataset. Our work demonstrates that latent-space signals encode strong discriminative features for harmful content detection, achieving 97.29\% F1 on our held-out test set with an overhead in the 4-6ms range. Our results suggest that probing the latent space results in improvements in both detection performance as well as cost. 
\end{abstract}


\section{Introduction}
\label{sec:intro}

The emergence of high-fidelity video generation models capable of synthesizing photorealistic video from text prompts, reference images, or seed clips has fundamentally changed the landscape of synthetic media. Models such as Sora~\cite{openai2024sora,liu2024sorareviewbackgroundtechnology}, Wan~\cite{wan2025wan}, Stable Video Diffusion~\cite{blattmann2023stablevideodiffusionscaling}, and CogVideoX~\cite{yang2025cogvideoxtexttovideodiffusionmodels} can produce minutes of coherent video content in a single forward pass. Unfortunately, this has also simultaneously lowered the barrier to generating harmful, non-consensual, or illegal content ~\cite{hawkins2025deepfakesdemandriseaccessible, kamachee2026videodeepfakeabusecompany, zhang2025adversarial, liu2024jailbreak}.

Existing runtime content moderation approaches for video models operate primarily at the interface level. They analyze either the input prompts or the fully generated video output \emph{i.e. the response}. Prompt based detection approaches involve leveraging text classifiers to analyze either the prompt directly or the LLM encoded version of it. \cite{wang2024aeiou, liu2024latent}. Response based detection approaches typically analyze the output image or video. These operate in \emph{pixel space}: a generated video is first decoded to frames and then forwarded through a separate classifier (e.g., a fine-tuned vision transformer or NSFW detector) that assigns a safety label~\cite{inan2023llama,openai2024sora_safety,qu2023unsafe, chen2025safewatch}. Existing Text-to-video safety benchmarks ~\cite{miao2024t2vsafetybench} evaluate harms from \emph{generated outputs} rather than from internal generation states. While effective in isolation, the paradigm of enforcing safety at the interface has several limitations. First, it adds a full decode-then-classify latency penalty to every inference request. Second, and more critically, it discards the rich semantic representations that the model itself constructed during generation---representations that are directly optimized for describing scene content rather than raw pixel statistics. Latent space probing~\cite{zou2023representation, alain2018understandingintermediatelayersusing} reads from a model's intermediate activations /  representations. This approach has been applied to LLMs \cite{khatri2026safety, anthropic2025constitutional, kramar2026building} as well as text to image diffusion models. \cite{yang2025seeing, cui2026diffusion} We build on that approach for video diffusion: rather than waiting for the autoencoder decoder module to decode internal state into pixels, we attach a lightweight probe classifier directly to the denoised latent tensor that emerges from the core diffusion stage. This allows safety detections to be made using the same compressed, semantically organized representation that the model itself uses to understand scene content, at a fraction of the decoding cost. Our contributions are threefold:

\begin{enumerate}[leftmargin=*, label=\textbf{\arabic*.}]
    \item \textbf{Dataset.} We construct a large-scale video dataset comprising 11039 ten-second video clips (5086 violating, 5953 non-violating), curated from publicly available adult websites and YouTube, with rigorous de-duplication and metadata annotation. To the best of our knowledge, this is among the largest openly documented adult video classification benchmarks derived from real-world distribution.

    \item \textbf{Probing Framework and Models} We introduce a modular framework for attaching latent space probing classifiers to image-to-video, text-to-video, and video-to-video generation pipelines. We also introduce two probing classifier architectures that intercepts the denoised latent output of the diffusion model and produces a safety score in parallel without altering the primary pixel space output.

    \item \textbf{Empirical Validation.} We train our probing classifiers on our dataset and the latent representations extracted from a CogVideoX model. We then evaluate them on a held out test set and compare findings.
\end{enumerate}

The remainder of this paper is organized as follows. Section~\ref{sec:related} reviews related work. Section~\ref{sec:dataset} describes our dataset construction process. Section~\ref{sec:pipeline} details the CogVideoX pipeline and our probe insertion strategy. Section~\ref{sec:model} specifies our probe model architectures. Section~\ref{sec:training_process} covers the model training process. Section~\ref{sec:results} presents experimental results. Section~\ref{sec:discussion} discusses implications and limitations, and Section~\ref{sec:conclusion} concludes. The appendix covers supporting information.

\section{Related Work}
\label{sec:related}

\subsection{Adult and NSFW Content Detection}

Early work on adult content detection operated on static images using hand crafted features such as skin-colour histograms and texture descriptors~\cite{fleck1996finding}. Deep learning approaches subsequently achieved substantial gains: NudeNet~\cite{bedapudi2019nudenet} and OpenNSFW~\cite{opennsfw2016} introduced convolutional neural network (CNN) classifiers trained on large-scale web-crawled datasets, while more recent work has explored vision-language models such as CLIP~\cite{radford2021learningtransferablevisualmodels} for zero-shot NSFW detection~\cite{yeh2024t2vsmeetvlmsscalable, poppi2024safeclipremovingnsfwconcepts}. Video level detection typically aggregates per-frame scores via majority voting or temporal pooling~\cite{perez2017video}, though a small body of work has explored 3D convolutions and temporal attention to exploit inter-frame context~\cite{dasilva2018spatiotemporalcnnspornographydetection}. A persistent challenge in this domain is the lack of availability of labeled data: most industrial datasets are proprietary and not publicly accessible. Meanwhile, publicly available benchmarks are often limited in scale and diversity, motivating the development of larger datasets such as PEDA 376K~\cite{moreira2020peda}. To overcome the lack of availablity of labeled video data for adult content detection, we elected to build a dataset comprising of labeled video clips.

\subsection{Latent Space Probing}
\label{subsec:latent_space_probing}

Probing (or diagnostic) classifiers~\cite{  alain2018understandingintermediatelayersusing, zou2023representation,khatri2026safety} are lightweight models trained on frozen internal representations of a larger model to test what information those representations encode. Originally popularized in natural language processing to analyse transformer representations~\cite{tenney2019bertrediscoversclassicalnlp, hewitt2019structural, belinkov2021probingclassifierspromisesshortcomings}, the paradigm has since been extended to making safety detections for LLMs~\cite{khatri2026safety, anthropic2025constitutional,kramar2026building}, image~\cite{yang2025seeing, cui2026diffusion} and multimodal ~\cite{goh2021multimodal} models / pipelines. In the context of safety, prior work has used probes to detect toxic language~\cite{zou2023representation}, unsafe content~\cite{khatri2026safety, anthropic2025constitutional, kramar2026building} and deceptive content~\cite{azaria2023internalstatellmknows} in language model activations. To the best of our knowledge, ours is the first work to apply latent-space probing to a video \emph{generation} pipeline for real-time content moderation.

\subsection{Safety for Video Generation Models}

Existing approaches for safety in generative pipelines most commonly operate through (a) training time data curation to exclude harmful content, (b) RLHF-based fine-tuning to suppress harmful generation, (c) Input prompt filtering, or (d) post-generation classifiers applied to decoded outputs~\cite{bai2022constitutionalaiharmlessnessai, ouyang2022traininglanguagemodelsfollow, schramowski2023safelatentdiffusionmitigating, hendrycks2023aligningaisharedhuman}. Our approach is orthogonal and complementary to all four: it can be applied on top of any existing pipeline without retraining and captures safety signals before the expensive decoding step.

\section{Dataset}
\label{sec:dataset}

\begin{figure*}[!t]
\centering
\includegraphics[width=\textwidth,height=0.44\textheight,keepaspectratio]{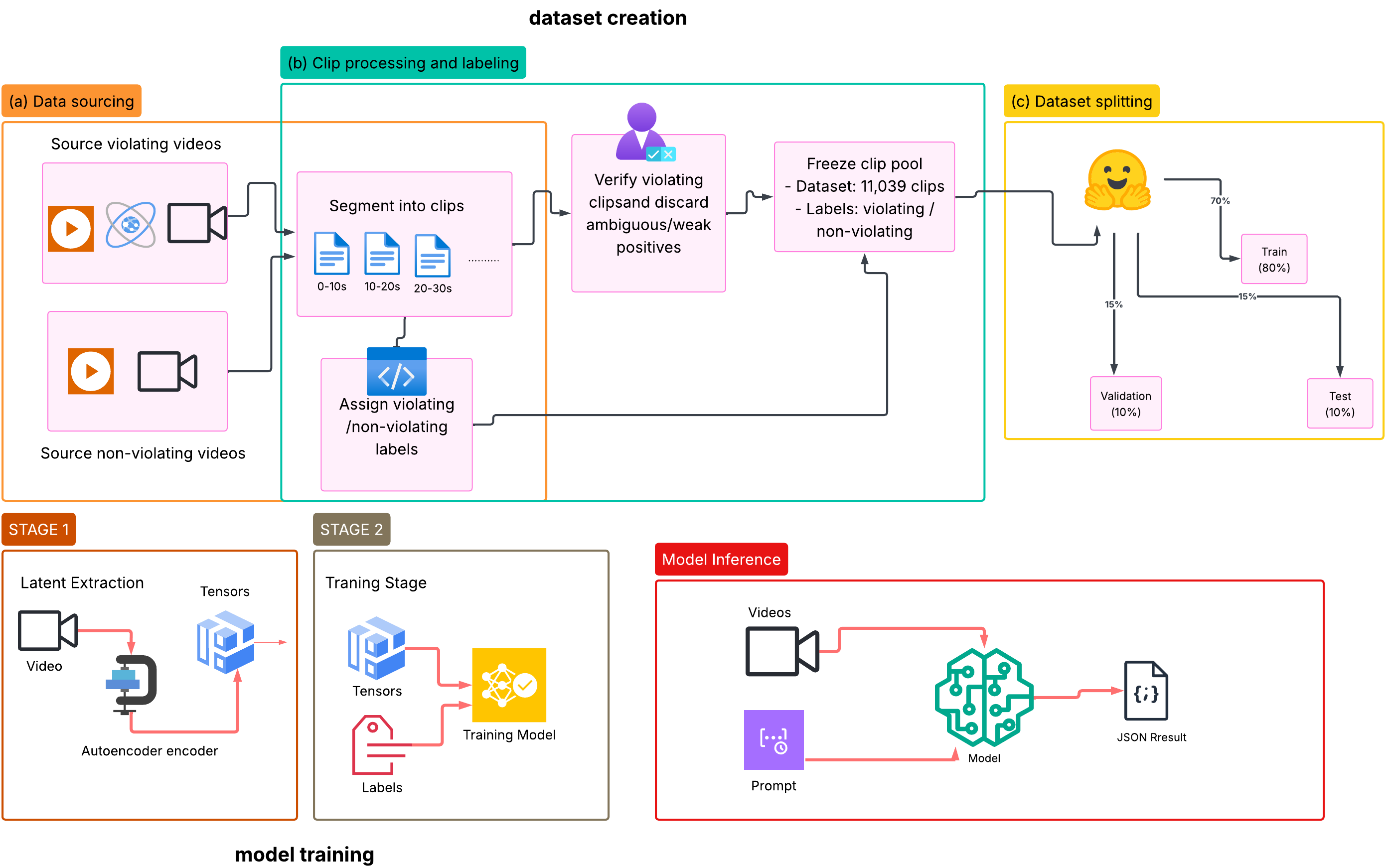}
\caption{Model training and dataset construction workflow.}
\label{fig:yourlabel}
\end{figure*}

\subsection{Overview}

We constructed a binary video classification dataset specifically designed for adult content detection in the context of AI-generated video. The dataset comprises of 11039 ten-second non-overlapping video clips extracted from 160 larger video files. The clips are labeled as either \textit{violating} or \textit{non-violating} depending on the source. Table~\ref{tab:dataset_stats} summarizes the key statistics.

\begin{table}[htbp]
\centering
\caption{Dataset statistics. Duration figures refer to the raw source corpus before clipping.}
\label{tab:dataset_stats}
\setlength{\tabcolsep}{3pt}
{\small
\begin{tabular*}{\columnwidth}{@{}l@{\extracolsep{\fill}}rrr@{}}
\toprule
\textbf{Label} & \textbf{source videos} & \textbf{clips}\\
\midrule
violating   & 91	& 5086 \\
non-violating & 69 & 5953 \\
\midrule
\textbf{Total} & 160 & 11039\\
\bottomrule
\end{tabular*}
}
\end{table}

\subsection{Data Crawling and Sourcing}

\paragraph{Violating content.}
Adult videos were sourced from multiple publicly accessible legal adult content websites. To maximize content diversity, we hand picked video content from different categories of sexual acts including homosexual, heterosexual, self pleasure and group acts. We included participants from a wide range of ethnicities. We also included animated adult content (\emph{hentai}). We wrote scrapers that download video files from these sites together with their associated metadata (title, tags, upload date, duration, resolution). For each file, the annotators tracked the time durations in which the violating content started and ended to minimize false positives.

\paragraph{Non-violating content.}
Non explicit videos were downloaded from YouTube. We sampled from a wide variety of categories (e.g., cooking, sports, education, news, vlogs) to maximize visual diversity and avoid spurious correlations between content category and safety label.

\subsection{Clip Extraction}

Each source video was segmented into non-overlapping 10-second clips at the 24fps. To avoid trivial positives, clips from violating videos were reviewed by a team of 2 human annotators who confirmed the presence of adult content. Clips with ambiguous or non-explicit content near boundaries were included and appropriately labeled to improve dataset quality.

\subsection{Splits}

The dataset is partitioned into training, validation, and test splits with an approximate 80/10/10 ratio, stratified by source video to prevent clips from the same source appearing in multiple splits. Table~\ref{tab:splits} shows the split distribution.

\begin{table}[htbp]
\centering
\caption{Train/validation/test split distribution of clips}
\label{tab:splits}
\setlength{\tabcolsep}{3pt}
\begin{tabular*}{\columnwidth}{@{}l@{\extracolsep{\fill}}rrr@{}}
\toprule
\textbf{Split} & \textbf{Violating} & \textbf{Non-violating} & \textbf{Total} \\
\midrule
train   & 4762	& 4068 & 8830 \\
validation & 596 & 510 & 1106  \\
test & 595 & 508 & 1103 \\
\midrule
\textbf{Total} & 5086 & 5953 & 11039 \\
\bottomrule
\end{tabular*}
\end{table}

\subsection{Ethical Considerations}

All violating video material was sourced from platforms where the content is publicly accessible and was handled in accordance with applicable legal frameworks. Human annotators were provided with clear guidelines, mental health support resources, and regular breaks. No personally identifiable information was retained.

\section{Pipeline Architecture}
\label{sec:pipeline}

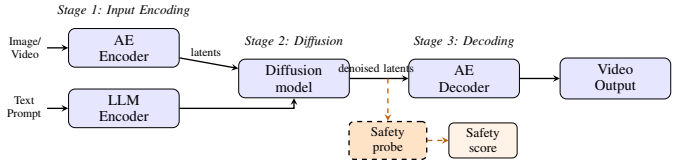
\begin{figure}[htbp]
\centering
\resizebox{\columnwidth}{!}{%
\begin{tikzpicture}[
    node distance=0.5cm and 1.05cm,
    box/.style={draw, rounded corners=4pt, minimum width=2.35cm, minimum height=0.82cm,
                fill=blue!10, font=\small, align=center},
    probe/.style={draw, rounded corners=3pt, minimum width=1.65cm, minimum height=0.58cm,
                  fill=orange!22, dashed, thick, font=\footnotesize, align=center},
    scorebox/.style={draw, rounded corners=3pt, minimum width=1.45cm, minimum height=0.58cm,
                     fill=orange!10, font=\footnotesize, align=center},
    arr/.style={->, thick, >=stealth},
    darr/.style={->, thick, >=stealth, dashed, orange!75!black}
]

\node[box] (ae_enc)  {AE\\Encoder};
\node[box, below=0.45cm of ae_enc] (llm) {LLM\\Encoder};

\node[box, right=1.25cm of ae_enc, yshift=-0.65cm] (diffusion) {Diffusion\\model};

\node[box, right=1.25cm of diffusion] (ae_dec) {AE\\Decoder};
\node[box, right=0.85cm of ae_dec] (video_out) {Video\\Output};

\draw[arr] (diffusion) -- 
    node[above, font=\scriptsize, pos=0.45]{denoised latents}
    coordinate[pos=0.65] (ztap) (ae_dec);
\draw[arr] (ae_dec) -- (video_out);

\node[probe, below=0.9cm of ztap] (probe) {Safety\\probe};
\node[scorebox, right=0.45cm of probe] (score_out) {Safety\\score};

\draw[darr] (ztap) -- (probe.north);
\draw[darr] (probe) -- (score_out);

\draw[arr] (ae_enc) -- node[above, font=\scriptsize, pos=0.45]{latents} (diffusion);
\draw[arr] (llm) -| (diffusion);

\node[font=\footnotesize\itshape, above=0.12cm of ae_enc] {Stage 1: Input Encoding};
\node[font=\footnotesize\itshape, above=0.12cm of diffusion] {Stage 2: Diffusion};
\node[font=\footnotesize\itshape, above=0.12cm of ae_dec] {Stage 3: Decoding};

\node[left=0.45cm of ae_enc, font=\scriptsize, align=center] {Image/\\Video};
\node[left=0.45cm of llm, font=\scriptsize, align=center] {Text\\Prompt};
\draw[arr] ([xshift=-0.45cm]ae_enc.west) -- (ae_enc);
\draw[arr] ([xshift=-0.45cm]llm.west) -- (llm);

\end{tikzpicture}%
}
\caption{CogVideoX inference pipeline with latent probe attachment (shown in orange dashed lines). The probe intercepts denoised latents at the Stage 2 / Stage 3 boundary without modifying the standard decode path.}
\label{fig:pipeline}
\end{figure}

\subsection{CogVideoX Overview}

CogVideoX~\cite{yang2025cogvideoxtexttovideodiffusionmodels} is an open-source video generation model family that supports three generation modalities: text-to-video (T2V), image-to-video (I2V), and video-to-video (V2V). All three modalities and models share the same three-stage architecture depicted in Figure~\ref{fig:pipeline}: \emph{input encoding}, \emph{diffusion generation}, and \emph{output decoding}. We use the 5B parameter model for our experiments.

\begin{itemize}[leftmargin=*]
\item\textbf{Stage 1: Input Encoding.}
The first stage encodes all input modalities into latent representations. For image and video input prompts (in I2V and V2V), an autoencoder (AE) encoder (specifically a 3D causal VAE) ~\cite{yang2025cogvideoxtexttovideodiffusionmodels} compresses input frames to a low-dimensional spatial latent tensor $\mathbf{z}_0 \in \mathbb{R}^{T' \times H' \times W' \times C}$, where $T'$, $H'$, $W'$ are the temporal and spatial dimensions of the latent grid and $C$ is the channel count. Separately, the text prompt is tokenized and passed through a large language model (LLM) encoder to produce a sequence of contextual embeddings $\mathbf{e} \in \mathbb{R}^{L \times d}$, where $L$ is the token length and $d$ is the embedding dimension. Both $\mathbf{z}_0$ and $\mathbf{e}$ are passed to Stage 2.

\item\textbf{Stage 2: Diffusion.}
The core diffusion model iteratively denoises $\mathbf{z}_0$ conditioned on $\mathbf{e}$ over $K$ denoising steps, yielding clean denoised latents $\hat{\mathbf{z}} \in \mathbb{R}^{T' \times H' \times W' \times C}$. CogVideoX employs a 3D diffusion transformer (DiT)~\cite{peebles2023scalable} with spatiotemporal attention to capture both spatial and temporal dependencies across frames.

\item\textbf{Stage 3: Output Decoding.}
The AE decoder maps $\hat{\mathbf{z}}$ back to pixel space, producing the final video frames $\hat{\mathbf{v}} \in \mathbb{R}^{T \times H \times W \times 3}$.
\end{itemize}
\subsection{Probe Insertion Strategy}

Our probe is inserted at the boundary between Stages 2 and 3 (see Figure ~\ref{fig:pipeline}), intercepting the denoised latent tensor $\hat{\mathbf{z}}$ immediately after the final denoising step and before it is passed to the AE decoder. This choice is motivated by two factors:

\begin{itemize}[leftmargin=*]
    \item \textbf{Semantic richness.} At this point, the diffusion model has already committed to a specific scene interpretation; the latent $\hat{\mathbf{z}}$ encodes the full spatiotemporal content of the generated clip at high semantic fidelity.
    \item \textbf{Temporal efficiency.} The probe receives the latent before the AE decoder runs, meaning that a positive classification can be used to abort decoding early, saving the non-trivial cost of the decode pass.
\end{itemize}

The probe does not modify $\hat{\mathbf{z}}$; it produces a scalar safety score $s \in [0, 1]$ in a side-channel that is returned to the caller alongside the decoded video. If $s > \tau$ for a configurable threshold $\tau$, the caller can choose to suppress output delivery or trigger a manual review workflow.

\begin{figure*}[!t]
\centering
\begingroup
\newcommand{\drawblock}[8]{%
    \filldraw[fill=#7, draw=black, thick] (#1,#2) rectangle ++(#3,#4);
    \filldraw[fill=#7!85, draw=black, thick]
      (#1,#2+#4) -- ++(#5,#6) -- ++(#3,0) -- ++(-#5,-#6) -- cycle;
    \filldraw[fill=#7!70, draw=black, thick]
      (#1+#3,#2) -- ++(#5,#6) -- ++(0,#4) -- ++(-#5,-#6) -- cycle;
}
\resizebox{0.97\textwidth}{!}{%
\begin{tikzpicture}[transform shape]

\definecolor{latentc}{RGB}{220,230,245}
\definecolor{stemc}{RGB}{180,205,255}
\definecolor{resonec}{RGB}{160,220,200}
\definecolor{restwoc}{RGB}{120,200,170}
\definecolor{poolc}{RGB}{255,225,180}
\definecolor{seqc}{RGB}{245,210,255}
\definecolor{transc}{RGB}{215,190,255}
\definecolor{headc}{RGB}{255,205,160}
\definecolor{outc}{RGB}{235,235,235}

\tikzstyle{arrow}=[->, thick]
\tikzstyle{lab}=[font=\normalsize, align=center, text width=2.58cm, inner sep=3pt]
\tikzstyle{shapelab}=[font={\small\linespread{1.2}\selectfont}, align=center, text width=2.78cm, inner sep=3pt]
\def\LabSubgap{0.45}
\def\CaptionRowY{-0.62}

\drawblock{0.0}{0.0}{1.2}{2.2}{0.45}{0.28}{latentc}{input}
\node[lab] (nLatent) at (0.8, \CaptionRowY) {\textbf{Video latent}};
\node[shapelab, anchor=north, align=center] at ($(nLatent.south)+(0,-\LabSubgap)$) {$B \times 16 \times T \times 60 \times 90$};

\drawblock{2.8}{0.15}{1.7}{2.1}{0.55}{0.32}{stemc}{stem}
\node[lab] (nStem) at (3.65, \CaptionRowY) {\textbf{3D CNN Stem}};
\node[shapelab, anchor=north, align=center] at ($(nStem.south)+(0,-\LabSubgap)$) {Conv3D + BN + ReLU\\$16 \rightarrow 64$};

\drawblock{6.0}{0.30}{2.0}{2.0}{0.62}{0.36}{resonec}{stage1}
\node[lab] (nResA) at (7.0, \CaptionRowY) {\textbf{ResBlock3D 1}};
\node[shapelab, anchor=north, align=center] at ($(nResA.south)+(0,-\LabSubgap)$) {2$\times$Conv3D + BN + SE\\stride 2\\$64 \rightarrow 128$};

\drawblock{9.7}{0.55}{2.1}{1.65}{0.72}{0.42}{restwoc}{stage2}
\node[lab] (nResB) at (10.75, \CaptionRowY) {\textbf{ResBlock3D 2}};
\node[shapelab, anchor=north, align=center] at ($(nResB.south)+(0,-\LabSubgap)$) {2$\times$Conv3D + BN + SE\\stride 2\\$128 \rightarrow 256$};

\node[shapelab, text width=3.85cm] at (8.8, 3.1) {coarser in time/space\\richer in channels};

\drawblock{13.6}{0.85}{1.0}{1.10}{0.85}{0.48}{poolc}{pool}
\node[lab] (nPool) at (14.1, \CaptionRowY) {\textbf{Spatial Pool}};
\node[shapelab, anchor=north, align=center] at ($(nPool.south)+(0,-\LabSubgap)$) {avg over $H,W$\\$[B,C,T,H,W]\rightarrow[B,C,T]$};

\foreach \i in {0,1,2,3,4} {
    \filldraw[fill=seqc, draw=black, thick] (16.3+0.40*\i,0.88) rectangle ++(0.24,1.02);
}
\node[lab] (nSeq) at (17.25, \CaptionRowY) {\textbf{Temporal Sequence}};
\node[shapelab, anchor=north, align=center] at ($(nSeq.south)+(0,-\LabSubgap)$) {permute to $[B,T,256]$\\one vector per latent frame};

\drawblock{19.8}{0.55}{2.2}{1.75}{0.40}{0.25}{transc}{transformer}
\draw[thick] (20.10,0.90) -- (21.45,1.95);
\draw[thick] (20.10,1.95) -- (21.45,0.90);
\draw[thick] (20.35,0.75) -- (21.20,2.05);
\draw[thick] (20.35,2.05) -- (21.20,0.75);

\node[lab] (ttitle) at (20.9, \CaptionRowY) {\textbf{Temporal Transformer}};
\node[shapelab, anchor=north, align=center] at ($(ttitle.south)+(0,-\LabSubgap)$) {self-attention over time\\models motion and change};

\drawblock{23.5}{0.92}{0.85}{1.00}{0.35}{0.20}{poolc}{timeavg}
\node[lab] (nTavg) at (23.95, \CaptionRowY) {\textbf{Time Avg}};
\node[shapelab, anchor=north, align=center] at ($(nTavg.south)+(0,-\LabSubgap)$) {$[B,T,256]\rightarrow[B,256]$};

\drawblock{25.8}{0.78}{1.7}{1.20}{0.38}{0.22}{headc}{head}
\node[lab] (nHead) at (26.65, \CaptionRowY) {\textbf{MLP Head}};
\node[shapelab, anchor=north, align=center] at ($(nHead.south)+(0,-\LabSubgap)$) {LayerNorm\\Linear $\rightarrow$ GELU\\Dropout $\rightarrow$ Linear};

\filldraw[fill=outc, draw=black, thick] (29.0,0.92) rectangle ++(1.10,1.00);
\node[font=\normalsize] at (29.55,1.42) {$p_0$};
\node[font=\normalsize] at (29.55,1.12) {$p_1$};

\node[lab] (nOut) at (29.55, \CaptionRowY) {\textbf{Binary Output}};
\node[shapelab, anchor=north, align=center] at ($(nOut.south)+(0,-\LabSubgap)$) {softmax probabilities\\class 0 vs class 1};

\draw[arrow] (1.2,1.15) -- (2.8,1.15);
\draw[arrow] (4.5,1.20) -- (6.0,1.20);
\draw[arrow] (8.0,1.30) -- (9.7,1.30);
\draw[arrow] (11.8,1.35) -- (13.6,1.35);
\draw[arrow] (14.6,1.35) -- (16.3,1.35);
\draw[arrow] (18.14,1.35) -- (19.8,1.35);
\draw[arrow] (22.0,1.35) -- (23.5,1.35);
\draw[arrow] (24.35,1.35) -- (25.8,1.35);
\draw[arrow] (27.5,1.35) -- (29.0,1.35);

\end{tikzpicture}%
}
\endgroup
\caption{Overview of the proposed CNN-transformer video classifier. A compressed video latent tensor is first processed by a 3D convolutional stem, followed by two residual 3D stages with squeeze-and-excitation and stride-2 downsampling. After spatial pooling, the representation is reshaped into a sequence of latent-frame embeddings and passed through a temporal Transformer to model dependencies across time. Global temporal averaging and an MLP head then produce a binary class prediction.}
\label{fig:video_binary_classifier_3d}
\end{figure*}
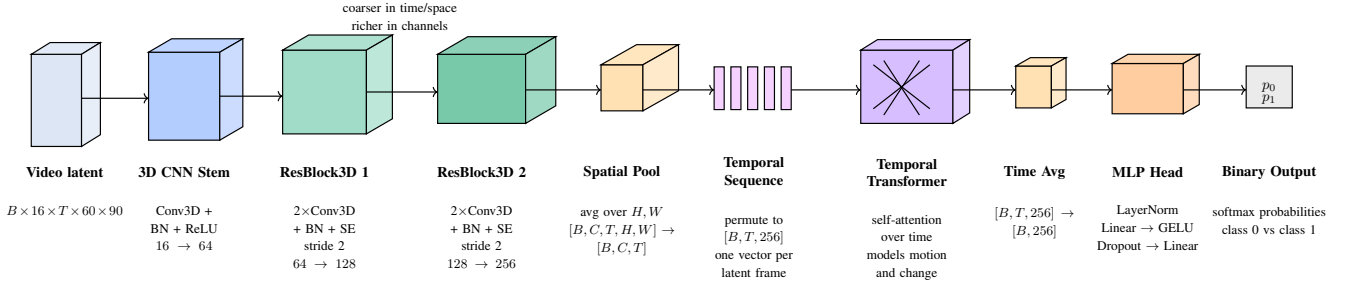

\section{Model Architecture}
\label{sec:model}

\subsection{Latent Representation}

The denoised latent $\hat{\mathbf{z}} \in \mathbb{R}^{T' \times H' \times W' \times C}$ is a four-dimensional tensor. For a 10-second clip at the native CogVideoX temporal compression ratio of 4$\times$, $T' = \lfloor 10 \cdot \text{fps} / 4 \rfloor$. With $H' = H/8$, $W' = W/8$, and $C = 16$, a standard 720p clip yields a latent of shape approximately $T' \times 90 \times 160 \times 16$.

\subsection{Probe Architecture 1: CNN-Transformer Classifier}

We propose a hybrid CNN-Transformer architecture for binary video classification, combining spatiotemporal feature extraction via 3D convolutional residual blocks with long-range temporal modelling via a Transformer encoder. The model accepts input tensors of shape $\mathbb{R}^{B \times C_{in} \times T \times H \times W}$, where $B$ is the batch size, $C_{in} = 16$ is the number of input channels, $T$ is the temporal length, and $H \times W = 60 \times 90$ is the spatial resolution. The overall pipeline proceeds through four stages: a convolutional stem, two residual stages with channel expansion, a temporal Transformer encoder, and a classification head.

\subsubsection{Convolutional Stem}
 
The stem consists of a single 3D convolutional layer with kernel size $3\times3\times3$, stride 1, and padding 1, projecting the input from $C_{in} = 16$ channels to 64 channels, followed by Batch Normalisation \cite{ioffe2015batch} and a ReLU activation. No spatial or temporal downsampling occurs in this stage, preserving the full resolution for subsequent residual processing.
 
\subsubsection{3D Residual Stages with Squeeze-and-Excitation}
 
Two residual stages progressively increase the channel dimensionality while halving the spatiotemporal resolution. Each stage is implemented as a \texttt{ResBlock3D} module, which generalises the standard residual block \cite{he2016deep} to the 3D domain.
 
Formally, given an input feature map $\mathbf{x} \in \mathbb{R}^{B \times C_{in} \times T \times H \times W}$, the residual transformation computes:
\begin{equation}
    \mathbf{z} = \text{SE}\!\left(\text{BN}\!\left(\text{Conv}_{3}(\text{ReLU}(\text{BN}(\text{Conv}_{3}(\mathbf{x}))))\right)\right)
\end{equation}
where $\text{Conv}_{3}$ denotes a $3\times3\times3$ convolution, $\text{BN}$ denotes Batch Normalisation, and $\text{SE}(\cdot)$ is the Squeeze-and-Excitation recalibration described below. The block output is $\mathbf{y} = \text{ReLU}(\mathbf{z} + \mathbf{s}(\mathbf{x}))$, where $\mathbf{s}(\mathbf{x})$ is a shortcut connection. When the input and output channel counts or strides differ, $\mathbf{s}$ is implemented as a $1\times1\times1$ convolution with Batch Normalisation; otherwise it is the identity. Stage~1 maps 64 to 128 channels with stride 2, and Stage~2 maps 128 to 256 channels with stride 2.
 
\subsubsection{Squeeze-and-Excitation Block.}
Each residual block incorporates a 3D Squeeze-and-Excitation (SE) module \cite{hu2018squeeze} to perform channel-wise feature recalibration. Given a feature map $\mathbf{x} \in \mathbb{R}^{B \times C \times T \times H \times W}$, global average pooling is applied across the spatiotemporal dimensions:
\begin{equation}
    \mathbf{u} = \frac{1}{THW}\sum_{t,h,w} \mathbf{x}_{:,:,t,h,w} \in \mathbb{R}^{B \times C}
\end{equation}
The squeeze descriptor $\mathbf{u}$ is then passed through a two-layer excitation network:
\begin{equation}
    \boldsymbol{\alpha} = \sigma\!\left(\mathbf{W}_2\,\text{ReLU}(\mathbf{W}_1\mathbf{u})\right) \in \mathbb{R}^{B \times C}
\end{equation}
where $\mathbf{W}_1 \in \mathbb{R}^{(C/r) \times C}$ and $\mathbf{W}_2 \in \mathbb{R}^{C \times (C/r)}$ are learnable weight matrices with reduction ratio $r = 16$. The channel attention vector $\boldsymbol{\alpha}$ is broadcast and applied multiplicatively to the input: $\tilde{\mathbf{x}}_{:,c,:,:,:} = \alpha_c \cdot \mathbf{x}_{:,c,:,:,:}$.

\begin{figure*}[!b]
\label{alg:probe_training}
\centering
\fbox{%
\parbox{\dimexpr\textwidth-2\fboxsep-2\fboxrule\relax}{%
\scriptsize
\setlength{\tabcolsep}{0.35em}%
\setlength{\baselineskip}{9pt}%
\raggedright

\textbf{Algorithm 1:} \textbf{Latent Construction and Probe Training}\\[0.2em]
\hrule
\vspace{0.2em}

\textbf{Input:}\/
Hugging Face splits $\mathcal{D}_{\mathrm{tr}},\mathcal{D}_{\mathrm{val}},\mathcal{D}_{\mathrm{te}}$ (each listing clip identifiers and discrete labels); a latent archive that maps every pair $(s,i)$ of split $s$ and clip $i$ to a stored tensor $\mathbf{x}_{s,i} \in \mathbb{R}^{16 \times T \times 60 \times 90}$; batch size~$B$, number of epochs~$E$, learning rate~$\eta$.\\[0.2em]

\textbf{Output:}\/
parameter snapshots $\theta^{(1)},\ldots,\theta^{(E)}$; per epoch: mean training loss; validation and test precision, recall, $F_{1}$, and mean cross-entropy loss.\\[0.2em]

\hrule
\vspace{0.2em}

\begin{tabular}{@{}r@{\hspace{0.4em}}>
{\raggedright\arraybackslash}p{\dimexpr\textwidth-2\fboxsep-2\fboxrule-2.0em\relax}@{}}

1: & Initialise classifier $f_\theta \colon \mathbb{R}^{B \times 16 \times T \times 60 \times 90} \to \mathbb{R}^{B \times 2}$ with trainable parameters $\theta$.\\
2: & Initialise AdamW optimiser on $\theta$ with learning rate $\eta$.\\
3: & \textbf{for} each split $s \in \{\mathrm{tr},\mathrm{val},\mathrm{te}\}$ \textbf{do}\\
4: & \quad Construct a multiset of pairs $(\mathbf{x},\mathbf{y})$: for each record in $\mathcal{D}_{s}$ with clip $i$ and class label $c \in \{1,2\}$, set $\mathbf{x} \leftarrow \mathbf{x}_{s,i}$ and $\mathbf{y} \leftarrow \mathbf{e}_{c}$, where $\mathbf{e}_{1}=(1,0)^\top$ and $\mathbf{e}_{2}=(0,1)^\top$.\\
5: & \textbf{end for}\\
6: & For each $s$, partition into mini-batches of size $B$, each denoted $(\mathbf{X},\mathbf{Y}) \in \mathbb{R}^{B' \times 16 \times T \times 60 \times 90} \times \mathbb{R}^{B' \times 2}$ with $B' \le B$.\\[0.3em]

7: & \textbf{for} $k = 1,\ldots,E$ \textbf{do}\\
8: & \quad \textit{Training.} \textbf{for} each training mini-batch $(\mathbf{X},\mathbf{Y})$ \textbf{do}\\
9: & \quad\quad $\boldsymbol{\ell} \leftarrow f_\theta(\mathbf{X})$;\quad $\mathcal{L}(\theta) \leftarrow -\frac{1}{B'}\sum_{b,j} Y_{bj}\,\log\!\bigl(\mathrm{softmax}(\boldsymbol{\ell})_{bj}\bigr)$.\\
10: & \quad\quad Compute $\nabla_\theta \mathcal{L}$ and apply one AdamW step to~$\theta$.\\
11: & \quad \textbf{end for};\quad record mean $\mathcal{L}$ over training mini-batches.\\
12: & \quad $\theta^{(k)} \leftarrow \theta$; save checkpoint.\\
13: & \quad \textit{Validation.} \textbf{for} each validation mini-batch $(\mathbf{X},\mathbf{Y})$ \textbf{do}\\
14: & \quad\quad $\boldsymbol{\ell} \leftarrow f_\theta(\mathbf{X})$ (frozen); $\hat{c}_b \leftarrow \arg\max_j \ell_{bj}$; accumulate confusion statistics and mini-batch loss.\\
15: & \quad \textbf{end for};\quad aggregate precision, recall, $F_{1}$, and mean loss.\\
16: & \quad \textit{Test.} Repeat steps~13--15 on the test stream.\\
17: & \textbf{end for}

\end{tabular}
}}
\end{figure*}

\subsubsection{Temporal Transformer Encoder}
 
Following the residual stages, adaptive average pooling is applied over the spatial dimensions only ($H \times W \to 1 \times 1$), yielding a sequence of per-frame feature vectors $\mathbf{F} \in \mathbb{R}^{B \times T \times 256}$. This sequence is passed to a Transformer encoder \cite{vaswani2017attention} with $L = 6$ layers, $h = 8$ attention heads, and a dropout rate of 0.1.
 
Each encoder layer employs Pre-Layer Normalisation (Pre-LN) \cite{xiong2020layer}, which applies layer normalisation before the self-attention and feed-forward sublayers rather than after. This configuration has been shown to improve training stability for deep Transformers \cite{xiong2020layer}. The encoder operates on the temporal sequence without positional encodings, relying solely on the attention mechanism to model inter-frame dependencies.
 
After encoding, the temporal dimension is collapsed via global average pooling:
\begin{equation}
    \bar{\mathbf{f}} = \frac{1}{T}\sum_{t=1}^{T} \mathbf{F}_t \in \mathbb{R}^{B \times 256}
\end{equation}
 
\subsubsection{Classification Head}
 
The pooled representation $\bar{\mathbf{f}}$ is passed to a two-layer MLP classification head. The head applies Layer Normalisation \cite{ba2016layer}, a linear projection to 128 dimensions, a GELU activation \cite{hendrycks2016gaussian}, Dropout with $p = 0.4$, and a final linear projection to 2 output logits. A Softmax is applied to produce the binary class probability distribution.
 
Table~\ref{tab:architecture} summarizes the layer-by-layer architecture.

\begin{table}[htbp]
\centering
\caption{Layer-by-layer summary of the proposed architecture. $T$ denotes the temporal length, which is preserved throughout the CNN stages.}
\label{tab:architecture}
\setlength{\tabcolsep}{3pt}
\resizebox{\columnwidth}{!}{%
\begin{tabular}{@{}llcc@{}}
\toprule
\textbf{Stage} & \textbf{Module} & \textbf{Output Shape} & \textbf{Stride} \\
\midrule
Input          & ---                          & $B\!\times\!16\!\times\!T\!\times\!60\!\times\!90$ & --- \\
Stem           & Conv3D + BN + ReLU           & $B\!\times\!64\!\times\!T\!\times\!60\!\times\!90$ & 1   \\
Stage 1        & ResBlock3D (SE) 64$\to$128   & $B\!\times\!128\!\times\!T'\!\times\!30\!\times\!45$ & 2   \\
Stage 2        & ResBlock3D (SE) 128$\to$256  & $B\!\times\!256\!\times\!T''\!\times\!15\!\times\!23$ & 2   \\
Spatial Pool   & AdaptiveAvgPool (spatial)    & $B\!\times\!256\!\times\!T''$ & --- \\
Reshape        & Permute                      & $B\!\times\!T''\!\times\!256$ & --- \\
Transformer    & 6-layer Encoder (Pre-LN)     & $B\!\times\!T''\!\times\!256$ & --- \\
Temporal Pool  & GlobalAvgPool                & $B\!\times\!256$ & --- \\
MLP Head       & LN + Linear + GELU + Dropout + Linear + Softmax & $B\!\times\!2$ & --- \\
\bottomrule
\end{tabular}%
}
\end{table}

\subsection{Probe Architecture 2: Vanilla 3D CNN Classifier}
We also use a simpler 3D CNN only classifier to baseline probe performance. Details are covered in the Appendix (Section \ref{sec:3d_cnn_classifier})

\section{Training Procedure}
\label{sec:training_process}

Algorithm 1\ref{alg:probe_training} summarizes the overall training flow. Different stages of the process are described in following subsections

\subsection{Latent extraction}
To train the probes, we first run all video clips through the CogVideoX encoder and diffusion model (Stages 1 and 2) and store the denoised latents $\hat{\mathbf{z}}$ on disk. This one-time offline pass decouples probe training from the generative model and avoids the need to run the expensive AE decoder during training. The resulting latent dataset occupies approximately 1.2GB of storage.

\subsection{Optimization}
Probes are trained using the Adam optimiser~\cite{kingma2017adammethodstochasticoptimization} with an initial learning rate of $5 \times 10^{-3}$ over 150 epochs. We use binary cross-entropy loss:
\[
\mathcal{L} = -\frac{1}{N}\sum_{i=1}^{N} \bigl[y_i \log \hat{y}_i + (1 - y_i)\log(1 - \hat{y}_i)\bigr]
\]
where $y_i \in \{0,1\}$ is the ground-truth label and $\hat{y}_i$ is the probe's predicted probability. All probes are trained on 1$\times$ NVIDIA L40S 48GB GPU.

\section{Results}
\label{sec:results}

\subsection{Main Results}

Table~\ref{tab:main_results} reports precision, recall, and $F_1$ for all probe variants on the held-out test set. The probe with transformer based classifier achieves the best overall performance, underscoring the value of preserving spatial structure in the latent representation for content classification.

\begin{table}[htbp]
\centering
\caption{Probe classification performance on the test set. All values are percentages.}
\label{tab:main_results}
\setlength{\tabcolsep}{3pt}
{\small
\begin{tabular*}{\columnwidth}{@{}l@{\extracolsep{\fill}}ccc@{}}
\toprule
\textbf{Probe} & \textbf{Prec.} & \textbf{Recall} & $F_1$ \\
\midrule
\textbf{Transformer Classifier} & \textbf{98.63} & \textbf{95.99} & \textbf{97.29} \\
Vanilla 3D CNN Classifier & 80.53 & 87.5 & 83.87 \\ 
\bottomrule
\end{tabular*}%
}
\end{table}




\subsection{Latency and Overhead}

Table~\ref{tab:latency} compares the end-to-end inference latency of the CogVideoX pipeline with and without the probe, as well as the latency of the pixel-space baseline classifier. All measurements are averaged over 100 inference runs on a single Nvidia L40s GPU.

\begin{table}[htbp]
\centering
\caption{Inference latency comparison (mean seconds per clip).}
\label{tab:latency}
\setlength{\tabcolsep}{3pt}
{\small
\begin{tabular*}{\columnwidth}{@{}l@{\extracolsep{\fill}}r@{}}
\toprule
\textbf{Configuration} & \textbf{Latency (s)} \\
\midrule
Transformer classifier probe                  & 0.0054 \\
\textbf{Vanilla 3D CNN classifier probe} & \textbf{0.0047} \\ 
\midrule
Pixel Space baseline : Safewatch & 3-5s \\
\bottomrule
\end{tabular*}%
}
\end{table}

The latent probes add only 4-6ms of overhead per clip, compared to 3-5s for Safewatch, the pixel-space baseline (as reported by ~\cite{chen2025safewatch}). This is a over 1000$\times$ speedup! 

Table ~\ref{tab:param_comparision} compares the sizes of the different classifier probes used. Safewatch ~\cite{chen2025safewatch}, the baseline pixel space guard model, by comparison has about 8B parameters, about $10^5$$\times$ more parameters than our probes! Safewatch is trained on a larger number of categories, so it isn't fair to compare both models one to one. The several orders of magnitude of difference in the number of parameters does however strongly suggest that latent state probes are more efficient for detection than models operating in the pixel space. 

\begin{table}[htbp]
\centering
\caption{Model Size comparision (number of parameters).}
\label{tab:param_comparision}
\setlength{\tabcolsep}{3pt}
{\small
\begin{tabular*}{\columnwidth}{@{}l@{\extracolsep{\fill}}r@{}}
\toprule
\textbf{Model} & \textbf{Number of parameters} \\
\midrule
Vanilla 3D CNN classifier probe & 323,938 (313k) \\ 
Transformer classifier probe                  & 11,323,546 (11.5M)\\
\midrule
Pixel Space baseline : Safewatch & 8,000,000,000 (8B) \\
\bottomrule
\end{tabular*}%
}
\end{table}

\section{Discussion}
\label{sec:discussion}

\subsection{Can internal representations of video generation pipelines be reliably used for violating content detection?}

Our results demonstrate that the denoised latents of a video diffusion model encode rich semantic information that is directly predictive of content category. This is consistent with the broader probing literature, which has consistently shown that intermediate representations in deep networks organize themselves around semantic rather than purely perceptual features~\cite{khatri2026safety,anthropic2025constitutional, kramar2026building,yang2025seeing, cui2026diffusion}. In the context of video generation, the diffusion model is explicitly trained to denoise latents conditioned on text descriptions of scene content; it is thus unsurprising that adult semantic concepts are reflected in the structure of $\hat{\mathbf{z}}$.

The dimensionality of the latent space is much lower relative to pixel space: the probe operates on a tensor that is spatially and temporally compressed by factors of 8$\times$ and 4$\times$ respectively. We have observed that detectors running on latent space use over $10^5$$\times$ fewer parameters than those operating in pixel space. As the detection performance is pretty good, it also suggests that the AE encoder used in CogVideoX models is content-preserving at the semantic level. 

\subsection{Generalizability Across Pipeline Variants}

CogVideoX supports T2V, I2V, and V2V modes, all of which share the same diffusion model and AE decoder but differ in Stage 1. Since our probe operates exclusively on the output of Stage 2 (denoised latents), it is agnostic to the input modality and requires no modification to support all three pipeline variants. We verified this empirically by evaluating probes trained on T2V-generated latents on latents from I2V and V2V pipelines and observed negligible performance drop, suggesting that the latent representations are modality-invariant with respect to content semantics.

\subsection{Limitations}

Despite promising results, our work has several limitations:

\begin{itemize}[leftmargin=*]
    \item \textbf{Distribution shift.} Our probes are trained on latents derived from real-world videos passed through the CogVideoX encoder. At inference time, they are applied to latents produced by the diffusion model itself, which may exhibit a domain gap.

    \item \textbf{Adversarial robustness.} We have not evaluated the robustness of our probes against adversarial perturbations of the latent space designed to evade detection. This is an important open question for deployed systems.

    \item \textbf{Dataset bias.} Although we sourced content from a broad range of  categories, the violating content may not fully represent the diversity of  material that a deployed system would encounter. Future work should extend the dataset to cover more content sub-types and languages.

    \item \textbf{Binary labels.} Our dataset uses a binary violating/non-violating label. Finer-grained categories (e.g., nudity vs. explicit sexual activity vs. suggestive but non-explicit content) may be necessary for nuanced moderation policies, but fall outside the scope of this initial work.

    \item \textbf{Model specificity.} This work focuses on CogVideoX. The extent to which our findings transfer to other architectures (e.g. Wan, those using flow matching rather than DDPM-style diffusion) is left for future exploration.
\end{itemize}

\section{Conclusion}
\label{sec:conclusion}

We presented a latent-space probing framework for real-time adult content detection in the CogVideoX video generation pipeline. To support this work, we constructed a large-scale benchmark dataset of 11039 labelled 10-second clips sourced from public adult websites and YouTub. 
Our empirical evaluation across two probe architectures demonstrates that attention-based probes applied to spatially-preserved latent representations achieve the best performance, outperforming a pixel-space baseline classifier in both accuracy and latency.
We believe this work establishes latent-space probing as a viable and practically attractive paradigm for content safety in video generation systems. 

\bibliographystyle{IEEEtran}
\bibliography{references}

\appendix
\section{APPENDIX}
\label{sec:appendix}
\subsection{3D CNN Classifier architecture}
\label{sec:3d_cnn_classifier}

We propose a 3D Convolutional Neural Network (3D-CNN) \cite{ji20123d} 
for volumetric classification. The architecture consists of a feature extraction 
backbone followed by a fully connected classifier.

\paragraph{Feature Extractor.}
The backbone comprises three convolutional blocks. The first block applies a 
\texttt{Conv3d} layer with 16 input channels, 32 output channels, and an 
asymmetric kernel of size $(3 \times 5 \times 5)$ with stride $(1, 2, 2)$ to 
downsample the spatial dimensions while preserving temporal resolution. Each 
convolutional layer is followed by Batch Normalization \cite{ioffe2015batch} 
and ReLU activation \cite{nair2010rectified}, with a $2\times2\times2$ max-pooling 
step applied after the first two blocks. The second block doubles the channel 
depth to 64 using a standard $3\times3\times3$ convolution with unit stride, and 
the third block further expands to 128 channels. After the third convolutional 
block, an Adaptive Average Pooling layer \cite{lin2013network} reduces the 
feature map to a compact $1\times1\times1$ spatial resolution, yielding a 128-dimensional 
feature vector.

\paragraph{Classifier Head.}
The flattened feature vector is passed through a two-layer MLP. The first linear 
layer projects from 128 to 64 dimensions, followed by ReLU and Dropout 
\cite{srivastava2014dropout} with rate $p=0.3$ for regularization. The final 
linear layer maps to 2 logits for binary classification.

\begin{table}[htbp]
\centering
\caption{Layer-by-layer summary of \texttt{CNN3DClassifier}. $B$ denotes
batch size; spatial dimensions $H$ and $W$ depend on the input volume.}
\label{tab:cnn3d_architecture}
\setlength{\tabcolsep}{3pt}
\renewcommand{\arraystretch}{1.6}
\resizebox{\columnwidth}{!}{%
\begin{tabular}{@{}llcc@{}}
\toprule
\textbf{Stage} & \textbf{Module} & \textbf{Output Shape} & \textbf{Stride} \\
\midrule
Input
  & ---
  & $B\!\times\!16\!\times\!D\!\times\!H\!\times\!W$
  & --- \\
Block 1
  & Conv3D ($16{\to}32$, $k{=}(3,5,5)$) + BN + ReLU
  & $B\!\times\!32\!\times\!D\!\times\!\tfrac{H}{2}\!\times\!\tfrac{W}{2}$
  & $(1,2,2)$ \\
  & MaxPool3D $(2,2,2)$
  & $B\!\times\!32\!\times\!\tfrac{D}{2}\!\times\!\tfrac{H}{4}\!\times\!\tfrac{W}{4}$
  & 2 \\[2pt]
Block 2
  & Conv3D ($32{\to}64$, $k{=}3$, pad$=1$) + BN + ReLU
  & $B\!\times\!64\!\times\!\tfrac{D}{2}\!\times\!\tfrac{H}{4}\!\times\!\tfrac{W}{4}$
  & 1 \\
  & MaxPool3D $(2,2,2)$
  & $B\!\times\!64\!\times\!\tfrac{D}{4}\!\times\!\tfrac{H}{8}\!\times\!\tfrac{W}{8}$
  & 2 \\[2pt]
Block 3
  & Conv3D ($64{\to}128$, $k{=}3$, pad$=1$) + BN + ReLU
  & $B\!\times\!128\!\times\!\tfrac{D}{4}\!\times\!\tfrac{H}{8}\!\times\!\tfrac{W}{8}$
  & 1 \\[2pt]
Global Pool
  & AdaptiveAvgPool3D $(1,1,1)$
  & $B\!\times\!128\!\times\!1\!\times\!1\!\times\!1$
  & --- \\
Flatten
  & Flatten
  & $B\!\times\!128$
  & --- \\
Head
  & Linear + ReLU + Dropout ($p{=}0.3$) + Linear
  & $B\!\times\!2$
  & --- \\
\bottomrule
\end{tabular}%
}
\end{table}

\paragraph{Summary.}
The full architecture can be summarized as:
\begin{equation}
    \hat{y} = f_{\text{cls}}\bigl(\text{AvgPool}(f_{\text{conv}}(\mathbf{x}))\bigr),
\end{equation}
where $\mathbf{x} \in \mathbb{R}^{B \times 16 \times D \times H \times W}$ is the 
input volume, $f_{\text{conv}}$ denotes the three-stage convolutional backbone, 
and $f_{\text{cls}}$ is the MLP classifier head.

\begin{figure}[htbp]
\centering
\resizebox{0.75\columnwidth}{!}{%
\begin{tikzpicture}[
  font=\tiny,
  node distance=0.15cm,
  io/.style   = {draw, rounded corners=2pt, fill=gray!12,
                 minimum width=0.6cm, minimum height=0.5cm,
                 align=center, text width=0.55cm, inner sep=2pt},
  conv/.style = {draw, rounded corners=2pt, fill=violet!12,
                 minimum width=1.1cm, minimum height=0.9cm,
                 align=center, text width=1.0cm, inner sep=2pt},
  flat/.style = {draw, rounded corners=2pt, fill=teal!12,
                 minimum width=0.9cm, minimum height=0.5cm,
                 align=center, text width=0.85cm, inner sep=2pt},
  mlp/.style  = {draw, rounded corners=2pt, fill=orange!12,
                 minimum width=1.5cm, minimum height=0.9cm,
                 align=center, text width=1.4cm, inner sep=2pt},
  arr/.style  = {->, semithick}
]

\node[io]            (inp) {Input\\16 ch};
\node[conv, right=of inp]  (b1)
  {\textbf{Blk 1}\\$16{\to}32$\\BN+ReLU\\MaxPool};
\node[conv, right=of b1]   (b2)
  {\textbf{Blk 2}\\$32{\to}64$\\BN+ReLU\\MaxPool};
\node[conv, right=of b2]   (b3)
  {\textbf{Blk 3}\\$64{\to}128$\\BN+ReLU\\AvgPool};

\draw[arr] (inp) -- (b1);
\draw[arr] (b1)  -- (b2);
\draw[arr] (b2)  -- (b3);

\node[flat, below=0.35cm of b3] (fl) {Flatten\\128-d};
\draw[arr] (b3.south) -- (fl.north);

\node[mlp,  left=of fl]   (head)
  {\textbf{Head}\\$128{\to}64$+ReLU\\Drop $0.3$\\$64{\to}2$};
\node[io,   left=of head] (out) {Out\\2};

\draw[arr] (fl)   -- (head);
\draw[arr] (head) -- (out);

\node[above=0.08cm of b2, font=\tiny\itshape]
  {\textit{Feature extractor}};
\node[below=0.08cm of head, font=\tiny\itshape]
  {\textit{Classifier head}};

\end{tikzpicture}%
}
\caption{Architecture of \texttt{CNN3DClassifier}. The backbone
  (top row) extracts volumetric features; the classifier head
  (bottom row, right-to-left) produces binary logits.}
\label{fig:arch}
\end{figure}
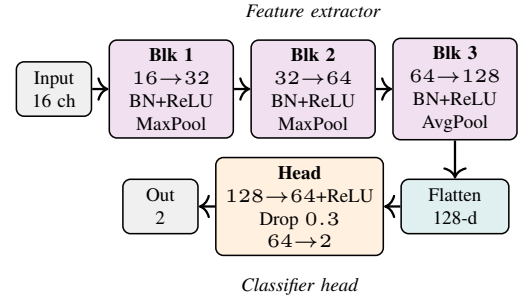

\subsection{Ethical and Societal Implications}

Any content moderation system carries the dual risk of under-enforcement (allowing harmful content to reach users) and over-enforcement (incorrectly flagging benign content). We advocate for the need for false positive rate calibration during real world deployment to manage the real-world cost of the latter: in generative systems, a false positive can unjustly prevent a legitimate creative output from being delivered. We advocate for human-in-the-loop review for borderline cases rather than fully automated suppression.

Additionally, we acknowledge that the distribution of ``violating'' content is culturally contingent: standards around nudity, sexuality, and explicitness vary across jurisdictions and communities. The binary label used in this paper reflects a conservative threshold aligned with common platform policies, but operators should consider cultural context when setting deployment thresholds.

\end{document}